# Leveraging Natural Learning Processing to Uncover Themes in Clinical Notes of Patients Admitted for Heart Failure

Ankita Agarwal[1], Krishnaprasad Thirunarayan[1], William L. Romine[1], Amanuel Alambo[1], Mia Cajita[2], and Tanvi Banerjee[1]

[1]Wright State University
[2]University of Illinois Chicago

*Abstract*— Heart failure occurs when the heart is not able to pump blood and oxygen to support other organs in the body as it should. Treatments include medications and sometimes hospitalization. Patients with heart failure can have both cardiovascular as well as non-cardiovascular comorbidities. Clinical notes of patients with heart failure can be analyzed to gain insight into the topics discussed in these notes and the major comorbidities in these patients. In this regard, we apply machine learning techniques, such as topic modeling, to identify the major themes found in the clinical notes specific to the procedures performed on 1,200 patients admitted for heart failure at the University of Illinois Hospital and Health Sciences System (UI Health). Topic modeling revealed five hidden themes in these clinical notes, including one related to heart disease comorbidities.

## I. INTRODUCTION

In the United States, about 6.2 million people have heart failure [1]. In 2014 alone, there were an estimated 3.4 million hospitalizations and 230,963 deaths with comorbid heart failure [2]. Khan et al. [3] identified and studied the trends in major comorbidities in heart failure (HF). Electronic health records (EHR) of patients from numerous hospital visits can be analyzed to uncover the themes discussed in these notes and to identify the comorbidities in these patients. This is important to determine their hospital discharge destinations and subsequent care. Automated analysis of clinical notes can be a challenging task. We propose natural language processing (NLP) and machine learning techniques to identify topics of these notes as well as the comorbidities associated with heart failure. One of the key challenges faced by NLP applied to medical text is the accurate detection and abstraction of the underlying medical concepts. After preprocessing the clinical notes, as part of natural language processing pipeline, we performed negation detection since certain entities in a clinical note can be negated and it is critical to recognize these negated phrases reliably. Then we identified the hidden themes in these notes using topic modeling implemented through Latent Dirichlet Allocation. These topics can be eventually used to determine appropriate discharge destination and effective care.

## II. RELATED WORK

Plati et al. [4] used clinical features, echocardiogram, and laboratory findings to diagnose chronic heart failure. They investigated the incremental value of each feature type and achieved an accuracy (91.23%), sensitivity (93.83%), and specificity (89.62%) to detect heart failure when features from all categories were utilized. Choi et al. [5] implemented an Artificial Intelligence-Clinical Decision Support System (AI-CDSS) using a hybrid expert-driven and machine-learning approach to detect heart failure. Their decision support system showed a high diagnostic accuracy for heart failure and so they concluded that AI techniques can be useful for the reliable and timely diagnosis of heart failure even in absence of heart failure specialists. Guo et al. [6] conducted a comprehensive review of the literature between January 2015 and August 2020 by a search of the PubMed library database using keywords 'machine learning heart failure' and 'deep learning heart failure'. They found that several machine learning models have been developed for heart failure diagnostic and outcome prediction using variables derived from EHR, demographics, medical, laboratory, and image. Similarly, Tripoliti et al. [7] presented the machine learning methodologies applied for the prediction of adverse events, like destabilizations, re-hospitalizations, and mortality.

While different machine learning techniques have been employed in the context of heart failure for different objectives, the task of uncovering latent themes in clinical notes of patients with heart failure which could help the clinical researchers to infer patient's severity of the disease, including related comorbidities, has not been explored. We explore the following research questions: 1. What are the major themes in the clinical notes of patients associated with heart failure? 2. Based on the themes identified, can HF severity be inferred? 3. What are the major cardiovascular comorbidities associated with patients admitted for heart failure?

Research reported in this publication was supported by NCCIH of the National Institutes of Health under award number 5R01AT010413S1. Any opinions, findings, and conclusions or recommendations expressed in this material are those of the authors and do not necessarily reflect the views of the NIH. The experimental procedures involving human subjects described in this paper were approved by the Institutional Review Board.
Corresponding author: Ankita Agarwal (agarwal.15@wright.edu)

## III. DATA COLLECTION AND PREPROCESSING

We analyze a dataset of 1,200 patients obtained through the University of Illinois (UI) Chicago Center for Clinical and Translational Science (CCTS) Biomedical Informatics Core. It comprises the EHR of patients admitted to UI Health between August 2016 to August 2021 who were 18 years or older during their first hospitalization. We focused on the procedures consisting of 15,966 clinical notes of 1,200 patients containing details of procedures performed on these patients along with the findings of these procedures and impressions as given by the consulting doctor. An example of the clinical note in our dataset is:

*'EXAM: Chest one view frontal 24.: sob*
*FINDINGS: Compared to the examination of August 20 significant interval resolution of previously noted increased pulmonary vascular markings associated with pulmonary edema. There are no new infiltrates. Residual linear density left base could be from the skull atelectasis.*
*IMPRESSION: Interval resolution of the findings of pulmonary edema with no new infiltrates.'*

In our dataset, not all clinical notes contained the 'IMPRESSION' attribute. For our analysis, we extracted the notes containing the 'IMPRESSION' attribute. For example, we extracted the above clinical note, and used the IMPRESSION '*Interval resolution of the findings of pulmonary edema with no new infiltrates*' in our analysis. In this manner, we were able to extract from 6,614 notes. Thereafter, we removed punctuation and numbers from these notes. After some manual analysis of these notes, we removed generic words like 'personalname, alphanumericid, examination, preliminary, impression, chest, view, change, final, available, evidence, see, right, stable, date, compare, interpretation, portable, lung, leave, report, process, review, resident, radiologist, attend, finding, electronic, and sign,' which are commonly found in EHRs, in the same manner as traditional stop words. As a result of this preprocessing, some notes became empty and so we were left with 6,347 notes for our analysis.

## IV. METHODS

We briefly discuss important components of the NLP pipeline we used to process the clinical notes.

### A. Negation Detection

Negation detection is a crucial part of processing clinical notes soundly. For example, a clinical note like '*Grossly normal chest with no pulmonary edema*' represents a normal impression while a note like '*Moderate cardiomegaly and pulmonary edema with right moderate pleural effusion*' represents a disease. We employed a negation detection tool to preserve the semantics [8]. Specifically, after detecting the negated phrases, we explicitly affixed 'no' followed by an underscore in front of that phrase. For example, negated 'acute cardiopulmonary process' was rendered 'no_acute_cardioplumonary_process'.

### B. Topic Modeling

To extract the hidden themes in the clinical notes, we used Latent Dirichlet Allocation (LDA) [9] with term frequency-inverse document frequency (TF-IDF) feature representation of the notes. To find the optimal number of topics in our dataset, we generated a coherence plot using the Coherence Model[1] and found that 5 would be the optimal number of topics to explain the features in the dataset. Thereafter, we fitted a 5-topic LDA model and labeled each clinical note with the dominant topic. To identify the theme for each topic, we looked at the representative notes which had the percent contribution of 0.80 or higher to the topic. The themes thus identified can be abstracted as about: (1) *mild to moderate severity, (2) findings related to the abdomen and chest, (3) insignificant and persistent findings, (4) cardiovascular-related findings,* and *(5) cardiopulmonary-related findings*.

### C. Linguistic analysis to identify comorbidities

Delimiting and automatically filtering the notes that best provide reliable comorbidities associated with heart failure could be a challenging task. To identify the common comorbidities associated with heart failure in our dataset, we performed linguistic analysis on the notes. We first represented the 6,347 clinical notes in TF-IDF format and selected the top features for these notes using chi-square analysis based on a multinomial logistic regression forward entry procedure (based on score test at 95% confidence level). Then, for each topic, we found the number of clinical notes containing each of the top 20 features and selected the topic which contained a large number of clinical notes containing these features. Based on this procedure, we found that Topic 4, *cardiovascular-related findings,* contained the maximum number of clinical notes having these top 20 features. So, we focused on this topic for further discussion on heart failure and comorbidities associated with it.

## V. RESULTS

We now summarize the results for key steps in the NLP pipeline.

### A. Negation Detection

After performing negation detection, we were able to identify phrases which indicated a negative impression for the procedure performed on the patient. For example, if we were able to identify the phrase like 'no focal consolidation' in a clinical note using the negation detection tool, we could infer that the patient's lungs are filled with air and not something else. So, negation detection helped us in making semantics explicit to improve soundness of analysis.

### B. Topic Modeling

Some of the representative clinical notes belonging to each of the five topics are shown in Table 1.

---

[1] https://radimrehurek.com/gensim/models/coherencemodel.html

**Table 1: Themes related to each topic, percentage of notes, unique keywords, and representative notes for each topic**

| Topic number and theme | % of notes | Unique keywords among top 10 keywords | Representative clinical note |
|---|---|---|---|
| *1. mild to moderate severity* | 19.3 | Stable_chest, bilateral, evidence, pleural_effusions, Mild_cardiomegaly | Chest: Heart is mildly enlarged. There is no evidence of pericardial effusion. Mild-to-moderate atherosclerotic calcifications are seen throughout the visualized thoracic aorta. Pulmonary artery is mildly enlarged measuring 3.4 cm in diameter. There is interval development of 3.5 x 1.0 cm peripherally enhancing pleural based mixed fluid and air collection within the posterior aspect of the right upper lobe with surrounding interlobular septal thickening. |
| *2. findings related to abdomen and chest* | 20.7 | stomach, cardiomegaly, Stable, NG_tube, pulmonary_vascular_congestion | 1. Findings suspicious for a proximal. partial small bowel obstruction. 2. Moderate right pleural effusion. 3. Cirrhotic liver morphology. 4. Moderate volume ascites. 5. Postsurgical changes of small bowel resection in the right lower quadrant. 6. Slight increase in size of left adrenal nodule favored to represent an adenoma. |
| *3. insignificant and stable findings* | 15.5 | No_significant, No_change, cardiomegaly, Stable, bilateral | HEAD:1. Noncontrast brain is negative for intracranial hemorrhage or depressed calvarial fracture.2. Chronic encephalomalacic changes involving the right frontal parietal lobes consistent with prior area of infarct. 3. Probable chronic small vessel ischemic disease. NECK: 1. Noncontrast cervical spine is negative for cervical fracture deformity. |
| *4. cardiovascular-related findings* | 22.1 | cardiomegaly, thoracic_spine, Arteriosclerosis, moderate, arthritic | The descending aorta is normal in caliber. The celiac axis. superior mesenteric artery and inferior mesenteric artery origins are patent. There is mild scattered atherosclerotic disease. including mild atherosclerotic disease of the SMA origin with mild luminal narrowing. Right and left renal arteries are patent. Mild atherosclerotic disease of the infrarenal aorta. common iliac and external iliac arteries without focal high-grade stenosis. |
| *5. cardio pulmonary-related findings* | 22.4 | pulmonary_edema, pneumonia, cardiomegaly, acute_cardiopulmonary_process, atelectasis | 1. Interval placement of a right-sided central venous catheter with the tip terminating in the mid. 2. Moderate cardiomegaly with a globular morphology which may be secondary to pericardial effusion or cardiomyopathy. 3. Pulmonary vascular congestion with coarsened interstitial lung markings. nonspecific but could represent early interstitial edema. 4. Age indeterminant compression deformity of T6. better assessed on the CTA examination from earlier today. |

*Note:* Each representative clinical note had the percent contribution of 0.80 or higher for their respective topic.

C. *Linguistic analysis to infer severity and to identify comorbidities*

After performing chi square analysis, we came up with top 20 words which are best at distinguishing between the five topics in our dataset. The chi-square statistic, degrees of freedom, and the number of clinical notes for each theme containing these words are shown in Table 2. This table shows that the word 'aorta' which is the main artery is predominantly found in Topic 4, i.e., *cardiovascular-related findings*. Additionally, words like 'arteriosclerosis' in this topic, i.e., hardening of arteries, provide further evidence to support that the notes are related to heart failure and its clinical presentation. The presence of words like 'arthritic, degenerative, pulmonary_congestive, and thoracic_spine, infiltrates, and pulmonary edema' led us to conclude that these were the most common cardiovascular-related findings in patients admitted with heart failure in our dataset. Comorbidities could not be identified from the keywords.

**Table 2: Top 20 words best at distinguishing between the topics along with their chi-square statistic, degrees of freedom and the number of clinical notes for each theme containing each of these top 20 words.**

| Top 20 words | mild to moderate severity | findings related to abdomen and chest | insignificant and persistent findings | Cardiovascular-related findings | Cardiopulmonary-related findings |
|---|---|---|---|---|---|
| no_significant ($\chi^2(4) = 194.5$) | 19 | 10 | **66** | 8 | 17 |
| no_change ($\chi^2(4) = 184.4$) | 7 | 1 | **57** | 7 | 8 |
| stomach ($\chi^2(4) = 131.3$) | 24 | **289** | 22 | 68 | 27 |
| thoracic_spine ($\chi^2(4) = 130.5$) | 27 | 92 | 9 | **396** | 8 |
| arteriosclerosis ($\chi^2(4) = 125.5$) | 23 | 75 | 7 | **379** | 9 |
| stable_chest ($\chi^2(4) = 124.4$) | **92** | 23 | 12 | 31 | 61 |
| arthritic ($\chi^2(4) = 120.5$) | 8 | 48 | 4 | **297** | 4 |
| pulmonary_congestive ($\chi^2(4) = 106.4$) | 5 | 29 | 3 | **226** | 5 |
| degenerative ($\chi^2(4) = 102.8$) | 52 | **178** | 41 | **439** | 36 |
| aorta ($\chi^2(4) = 97.3$) | 34 | 78 | 20 | **383** | 17 |
| no_focal_consolidation ($\chi^2(4) = 79.8$) | 1 | 0 | 1 | 1 | **53** |
| infiltrates ($\chi^2(4) = 78.9$) | 60 | 77 | 17 | **214** | 38 |
| positions ($\chi^2(4) = 72.5$) | 11 | 27 | 0 | **118** | 1 |
| acute_cardiopulmonary_process ($\chi^2(4) = 66.3$) | 33 | 3 | 15 | 8 | **56** |
| ng_tubes ($\chi^2(4) = 62.9$) | 10 | 4 | 0 | 95 | 6 |
| elongated ($\chi^2(4) = 58.4$) | 3 | 15 | 1 | 88 | 0 |
| persistent_pulmonary_edema ($\chi^2(4) = 57.2$) | 1 | 0 | 18 | 0 | 0 |
| pulmonary_edema ($\chi^2(4) = 51.2$) | **125** | 84 | **128** | 211 | **291** |
| moderate ($\chi^2(4) = 49.4$) | **161** | **200** | 148 | **402** | **272** |
| acute_intrathoracic_process ($\chi^2(4) = 49.4$) | 7 | 1 | 0 | 3 | 44 |

*Note:* All words are significant at 0.01 significance level. For each topic, a high number of clinical notes for a particular feature are highlighted.

## VI. DISCUSSION

The five topics identified in our dataset uncovered the major themes found in the clinical notes of patients admitted for heart failure. The theme, *mild to moderate severity,* contained the notes where the patients' condition was mild to moderate. The theme, *findings related to abdomen and chest* contained notes about impressions on patients' abdomen and chest area and the word 'stomach' was found in a large number of these notes. The theme, *insignificant and persistent findings,* talked about the stable or no significant changes to the patient's condition. The theme, *cardiovascular-related findings,* talked about a person's heart condition as well as their pelvic and abdominal vasculature. For example, the impression in this topic: *'The descending aorta is normal in caliber. The celiac axis. superior mesenteric artery and inferior mesenteric artery origins are patent. There is mild scattered atherosclerotic disease.'* depicted the condition of the patient's vasculature. The last theme, *cardiopulmonary-related findings,* contained notes related to pulmonary diseases like edema, consolidation of lungs, and embolism. For example, the impression in this topic: *'Bilateral pleural effusions. right greater than left. which appear increased on the right and decreased on the left from prior study. Consolidation of the right midlung and lower lung zones also appear increased. Correlate with clinical factors for pneumonia'* revealed findings about the extent pulmonary involvement in a patient with heart failure.

Common heart failure comorbidities (for example, ischemic heart disease, diabetes, chronic kidney disease) were not listed among the top keywords; this could be because the analysis focused on procedural notes and comorbidities are typically noted in the patient's history and physical and discharge summary.

The clinical notes had a longitudinal set of notes for each patient. A number of procedures were conducted on a patient on different body parts like chest, abdomen, pelvis, spine, legs, hand, head, and shoulder, and at different times to determine the severity of the disease. A patient may be diagnosed to have a mild disease for a procedure performed on one body part while may have a severe disease in another body part. For example, in our dataset, a patient had a procedure being performed on chest and the impression, *'Constellation of findings can be seen with pulmonary edema. Superimposed multifocal pneumonia may also be considered'*. On the same patient, a procedure was performed on the head with the impression, *'Negative for acute intracranial hemorrhage and hydrocephalus'*. So, this patient has probable pneumonia but had no intracranial abnormalities. Thus, different themes found in the clinical notes for the same patient can be analyzed together to determine the severity of the disease, plan the course of action and treatment including hospital discharge. Automated classification of these notes into different topics for a patient can also be analyzed over time through a temporal representation to determine how the topic changes over time, such as progression from severe to stable disease, stable to severe disease, or emergence/disappearance of comorbidities. This is especially important given that heart failure is a chronic progressive disease.

In future, we intend to implement dynamic topic models to visualize how a specific clinical term changes its meaning over time to better understand progression of a patient's condition. For example, if a word like 'kidneys/renal' belonged to theme, *mild to moderate severity* or to theme, *insignificant and stable findings*, that could imply that the patient's heart failure has not progressed to the point where it's causing injury to the kidneys. On the other hand, if this word belonged to theme, *cardiovascular-related findings* it could indicate the patient's worsening cardiac function has led to acute kidney injury.

## REFERENCES

[1] S.S. Virani, A. Alonso, E.J. Benjamin, M.S. Bittencourt, C.W. Callaway, A.P. Carson, A.M. Chamberlain, A.R. Chang, S. Cheng, F.N. Delling, and L. Djousse, "Heart disease and stroke statistics—2020 update: a report from the American Heart Association. *Circulation*", *141*(9), pp. e139-e596, 2020.

[2] S. L. Jackson, X. Tong, R. J. King, F. Loustalot, Y. Hong, and M. D. Ritchey, "National Burden of Heart Failure Events in the United States, 2006 to 2014," *Circ. Heart Fail.*, vol. 11, no. 12, pp. e004873, Dec. 2018.

[3] M.S. Khan, A. Samman Tahhan, M. Vaduganathan, S.J. Greene, A. Alrohaibani, S.D. Anker, O. Vardeny, G.C. Fonarow, and J. Butler, "Trends in prevalence of comorbidities in heart failure clinical trials", *European journal of heart failure*, vol. 22, no. 6, pp.1032-1042, 2020.

[4] D.K. Plati, E.E. Tripoliti, A. Bechlioulis, A. Rammos, I. Dimou, L. Lakkas, C. Watson, K. McDonald, M. Ledwidge, R. Pharithi, and J. Gallagher, "A Machine Learning Approach for Chronic Heart Failure Diagnosis", *Diagnostics*, vol. *11, no.* 10, pp.1863, 2021.

[5] D.J. Choi, J. J. Park, T. Ali, and S. Lee, "Artificial intelligence for the diagnosis of heart failure," *NPJ Digital Medicine*, vol. 3, no. 1. pp. 1-6, 2020.

[6] A. Guo, M. Pasque, F. Loh, D.L. Mann, and P.R. Payne, "Heart failure diagnosis, readmission, and mortality prediction using machine learning and artificial intelligence models", *Current Epidemiology Reports*, pp.1-8, 2020.

[7] E. E. Tripoliti, T. G. Papadopoulos, G. S. Karanasiou, K. K. Naka, and D. I. Fotiadis, "Heart Failure: Diagnosis, Severity Estimation and Prediction of Adverse Events Through Machine Learning Techniques," *Computational and structural biotechnology journal*, vol. 15, pp. 26–47, 2017.

[8] W.W. Chapman, W. Bridewell, P. Hanbury, G.F. Cooper, and B.G. Buchanan, "A Simple Algorithm for Identifying Negated Findings and Diseases in Discharge Summaries," *Journal of Biomedical informatics*, vol. 34, no. 5, pp. 301–310, Oct. 2001.

[9] D. M. Blei, A. Y. Ng, and M. I. Jordan, "Latent dirichlet allocation," Journal of machine Learning research, vol. 3, no. Jan, pp. 993–1022, 2003.